\documentclass[11pt]{article}

\usepackage[utf8]{inputenc}

\usepackage{graphicx} 

\usepackage{ACL2023}

\usepackage{booktabs} 
\usepackage{times}
\usepackage{latexsym}
\usepackage{graphicx}
\usepackage{multirow}

\usepackage[T1]{fontenc}
\usepackage{float}
\usepackage{afterpage}
\usepackage{caption}
\usepackage[utf8]{inputenc}

\usepackage{microtype}

\usepackage{inconsolata}
\title{FSM: A Finite State Machine Based  Zero-Shot Prompting Paradigm \\ for Multi-Hop Question Answering}

\author{Xiaochen Wang$^{1, 2,3}$, Junqing He$^{3}$\footnotemark[1], Zhe Yang$^{1}$, Yiru Wang$^{4}$
  \\ 
\textbf{Xiangdi Meng}$^{1, 2}$, \textbf{Kunhao Pan}$^{3}$, \textbf{Zhifang Sui}$^{1}$\footnotemark[1] \\
$^1$National Key Laboratory for Multimedia Information Processing, \\
School of Computer Science, Peking University \\ 
$^2$School of Software \& Microelectronics, Peking University\\
$^3$International Digital Economy Academy 
$^4$ModelTC\\
\texttt{wangxiaochen@stu.pku.edu.cn } \texttt{szf@pku.edu.cn } \texttt{hejunqing@idea.edu.cn}\\
}

\begin{document}
\maketitle

\begin{abstract}

Large Language Models (LLMs) with chain-of-thought (COT) prompting have demonstrated impressive abilities on simple nature language inference tasks. However, they tend to perform poorly on Multi-hop Question Answering (MHQA) tasks due to several challenges, including hallucination, error propagation and 
limited context length. We propose a prompting method, Finite State Machine (FSM) to enhance the reasoning capabilities of LLM for complex tasks in addition to improved effectiveness and trustworthiness. 
Different from COT methods, FSM addresses MHQA by iteratively decomposing a question into multi-turn sub-questions, and self-correcting in time, improving the accuracy of answers in each step. Specifically, FSM addresses one sub-question at a time and decides on the next step based on its current result and state, in an automaton-like format. Experiments on benchmarks show the effectiveness of our method. Although our method performs on par with the baseline on relatively simpler datasets, it excels on challenging datasets like Musique. Moreover, this approach mitigates the hallucination phenomenon, wherein the correct final answer can be recovered despite errors in intermediate reasoning. Furthermore, our method improves LLMs' ability to follow specified output format requirements, significantly reducing the difficulty of answer interpretation and the need for reformatting.

\end{abstract}
\footnotetext[1]{Corresponding author.}
\section{Introduction}
Multi-hop Question Answering has intrigued researchers for its complexity and practical implications. 
Researchers employ two primary strategies to address MHQA using Large Language Models. One effective method is In-Context Learning (ICL) \cite{self-prompted,least_to_most}, where models are guided to solve problems based on detailed instructions, often through examples of problem decomposition.  However, few-shot methods with manual demonstrations are expansive and time-consuming. Another approach involves fine-tuning LLMs with domain-specific data, a complex process \cite{train_kuaishou} requiring substantial high-quality data and computational resources. This approach is unable to generalize to unseen datasets and domains without training.
Despite advancements in single-hop question answering, MHQA remains challenging due to the need to extract information from lengthy texts and conduct multi-step reasoning without supervision, which poses difficulties for LLMs. LLMs struggle with reading long texts and multi-step reasoning tasks.
\begin{figure}[!t]
    \centering
    \includegraphics[width=0.45\textwidth]{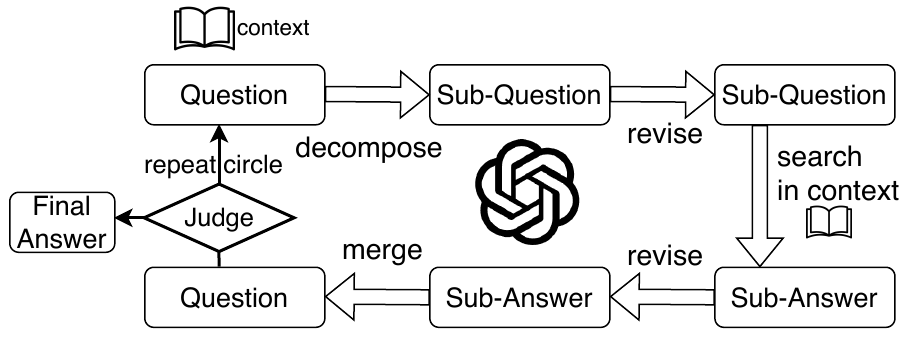}
    \caption{The abstract flow chart of FSM}
    \label{fig:FSM}
\end{figure}

\begin{figure*}[ht]
  \centering
  \includegraphics[ width=\textwidth]{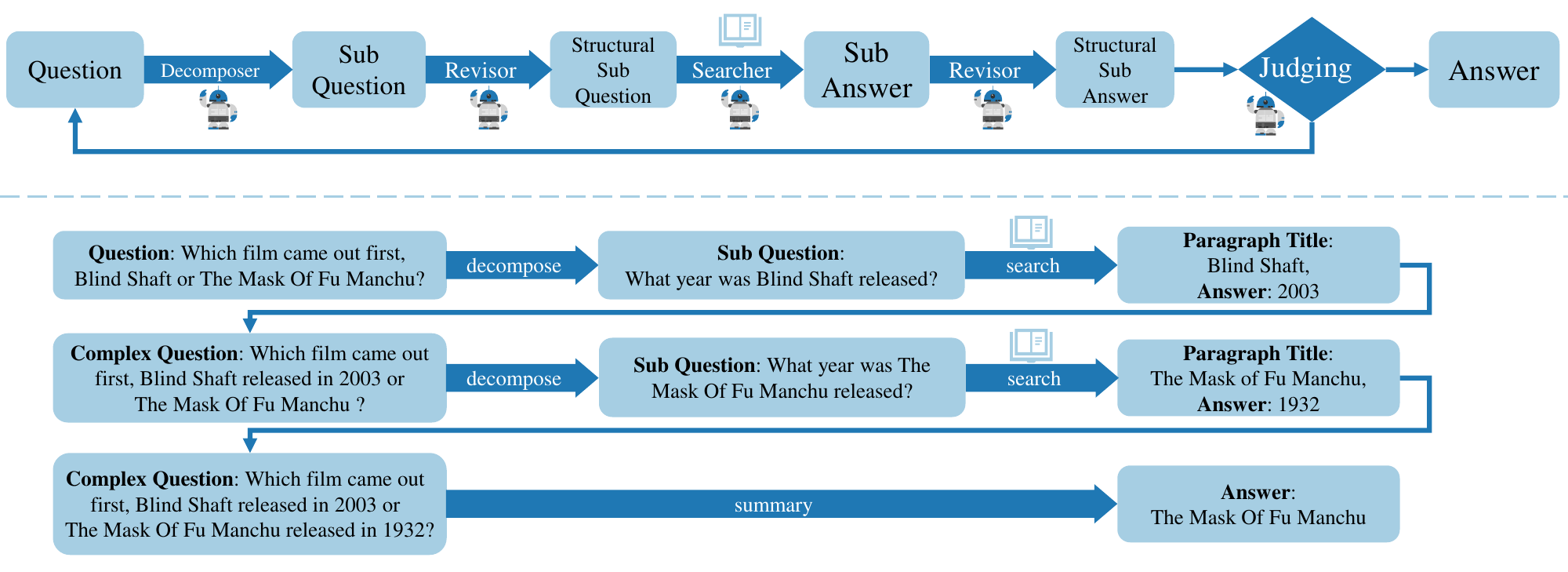}
  \caption{ The flow chart of proposed FSM and a simple case in detail. A multi-hop question is sovled step by step orderly. The book icon indicates candidate paragraphs in the search step. The robot denotes LLMs.}
  \label{fig_automat}
\end{figure*}

Why do LLMs underperform in current MHQA methods? By analyzing errors in existing approaches, we identified four common error types, which will be detailed in Section \ref{Sec:Disscusion}. Specific incorrect examples from common methods are illustrated in Figure \ref{fig_error}. We found that LLMs struggle particularly in intermediate reasoning stages, where errors in initial steps can propagate, leading to incorrect conclusions. Additionally, few-shot techniques like REACT \cite{react} and SP-COT \cite{self-prompted} need a minimum of 4-shot displays with long context, surpassing context boundaries.

According to the analysis above, we propose a zero-shot method named \textbf{F}inite \textbf{S}tate \textbf{M}achine prompting (FSM), simplifying the MHQA task into four sub-tasks: decomposing questions, searching for answers in candidate paragraphs, revising the format, judging whether to continue or summarizing with all key information. Figure \ref{fig_automat} depicts the process of the FSM. LLMs address one sub-question per round, deciding the next step based on the current state, following an automaton-like process. Clear and explicit sub-tasks, along with timely revisions, make the reasoning process more manageable and accurate.

Extensive experiments on MHQA benchmarks \cite{hotpot,musi,2wiki} demonstrate that our approach outperforms GPT and 72B LLM baselines, nearly doubling the F1 score on Musique \cite{musi}. Furthermore, unlike our framework, baselines have a high frequency of producing outputs in unexpected formats and type errors that require additional processing to extract correct answers.

Our contributions are as follows:

\textbullet~To address reasoning challenges in LLMs for MHQA tasks, we introduce FSM, a zero-shot prompting paradigm based on finite state machines to decompose complex questions iteratively. This approach aims to strengthen control over intermediate reasoning and improve overall accuracy.

\textbullet~ We investigate the reason for errors in MHQA and conduct various experiments on the insights. e.g. hallucination exists in direct answer predictions, and the contextual length is a bottleneck for reasoning. 

\textbullet~Extensive experiments on MHQA benchmarks in different settings validate FSM's effectiveness, especially on challenging datasets. The method can also be adapted to other complex tasks such as nature language to SQL (NL2SQL).


\section{Methodology}

\subsection{Strategies}
LLMs are weak in following instructions. Through manual observation of error examples \ref{fig_error}, we infer from the results of baseline methods that the model struggles with completing complex instructions in a single step. LLMs tend to forget previous instructions during reasoning. 

To address these issues, we propose the following strategies to reduce burden for LLMs: 

a) \textbf{Iterative Decomposition}: 
 
Unlike few-shot reasoning approaches, FSM adopts a multi-turn process. 
Each iteration focuses on addressing a single sub-task, enabling LLMs to understand instructions clearly and execute them accurately.

b) \textbf{Error Checking and Backtracking}: For each reasoning step, FSM conducts a verification check to ensure the correctness of response. If an irregular or incorrect output is identified, the model is allowed to self-revise the answer or backtrack.

c) \textbf{Final Review Step}: To minimize distractions from lengthy contexts, we utilize sub-questions, corresponding supporting factual paragraphs, evidence, and answers to further verify the consistency of answers and reasoning, named FSM2.

\subsection{Framework}

We present our proposed Finite State Machine (FSM) in two distinct stages as illustrated in Figure~\ref{fig_automat}. Initially, we instruct LLMs to address sub-questions iteratively during the first phase. Subsequently, in stage 2, LLMs are tasked with summarizing the responses incorporating key information from each sub-question. The FSM framework is depicted in Figure~\ref{fig_automat}.

To elaborate, our approach commences by assisting the model in breaking down the primary question into smaller components. Following this, we compare the original question with the sub-questions to ensure semantic equivalence; any disparities prompt the model to further decompose the elements. In the third phase, the model scans the context for related paragraphs, retrieving relevant information and answers. The fourth step entails revising the complex question with the response to the sub-question and identifying the relationship with updated complex question and sub-question, composition or comparison. Additionally, we conduct checks to ascertain whether the answer constitutes a simple or compound sentence, then promptly breaks down compound sentences. This iterative process continues until the revised question reaches a point where no further decomposition is feasible. By meticulously following each step, our methodology enables a more accurate evaluation of a model's true capabilities, distinguishing it from other approaches that tend to overlook crucial intermediate stages, which may yield seemingly correct outcomes despite flawed reasoning processes. We have included prompts for the whole process in the Appendix ~\ref{prompt}.

\begin{table*}[ht]
\centering

\resizebox{\textwidth}{!}{
\begin{tabular}{*{19}{c}}
\toprule[1pt]
 & \multirow{3}{*}{}  & \multicolumn{3}{c}{Musique} & \multicolumn{7}{c}{HotpotQA} & \multicolumn{7}{c}{2wiki} \\	
 \cmidrule(lr){3-5}\cmidrule(lr){6-12}\cmidrule(lr){13-19} 
 & & \multicolumn{3}{c}{ans}  & \multicolumn{2}{c}{ans} & \multicolumn{2}{c}{sup} & \multicolumn{3}{c}{joint} 
 & \multicolumn{2}{c}{ans} & \multicolumn{2}{c}{sup} & \multicolumn{3}{c}{joint}   \\ 
 
 \cmidrule(lr){3-5}\cmidrule(lr){6-7} \cmidrule(lr){8-9} \cmidrule(lr){10-12} \cmidrule(lr){13-14} \cmidrule(lr){15-16} \cmidrule(lr){17-19}

~ &  & EM & F1  & Format & EM & F1 & EM & F1 & EM & F1 & Format &  EM & F1 & EM & F1 & EM & F1 & Format \\
\midrule
 \multirow{4}{*}{\rotatebox[origin=c]{90}{Qwen}}  & \textbf{Normal} & 18.2 & 30.9  & 84.0 & 31.6 & 42.8 & \textbf{2.6} & 26.4 & \textbf{1.3} & 13.4 & 90.7 & 6.7 & 8.0 & 1.6 & 5.5 & 1.0 & \textbf{2.6} & 89.8  \\ 
& \textbf{COT} & 1.0 & 6.6  & 7.0 & 3.1 & 9.7 & 0.1 & 0.7 & 0.1 & 0.4 & 4.4 & 0.6 & 1.9 & 0 & 0.1 & 0.0 & 0.0 & 4.2   \\ 

& \textbf{FSM1} & \textbf{26.2} & \textbf{41.2}  & 100.0  & 22.5 & 33.3 & 0.7 & 9.9 & 0.4 & 3.6 & 100.0 & 27.6 & 37.9 & 4.7 & 25.8 & 1.9 & 9.1 & 100.0 \\ 
& \textbf{FSM2} & 21.9 & 37.7  & 100.0 & \textbf{33.1} & \textbf{46.0} & 1.8 & \textbf{28.8} & 1.0 & \textbf{15.7} & 100.0 & \textbf{36.1} & \textbf{49.3} & \textbf{7.7}& \textbf{38.4} & \textbf{5.1} & \textbf{19.4} & 100.0  \\ 

\midrule

\multirow{4}{*}{\rotatebox[origin=c]{90}{GPT}} & \textbf{Normal}  & 16.7 & 27.8 & 94.0 & 34.0 & 45.9 & 0.7 & 15.0 & 3.0 & 8.0 & 94.3 & \textbf{37.3} & \textbf{46.6} & 1.0 & 14.1 & \textbf{9.0} & 7.2 & 95.8 \\
& \textbf{COT} & 4.5 & 13.6 & 14.7 & 12.3 & 26.0 & 0.4 & 4.5 & 2.0 & 17.8 & 16.2 & 8.2 & 19.3 & 0.2 & 1.3 & 1.0 & 4.6 & 7.0  \\

& \textbf{FSM1}  & \textbf{26.0} & \textbf{38.4} & 100.0 & 23.4 & 32.0 & \textbf{2.4} & \textbf{29.3} & 2.0 & 9.8 & 100.0 & 30.1 & 40.0 & \textbf{14.2} & \textbf{47.0} & 2.0 & 8.5 & 100.0 \\
& \textbf{FSM2} & 18.6 & 27.4  & 100.0 & \textbf{28.4} & \textbf{36.7} & 2.2 & 21.4 & \textbf{4.0} & \textbf{26.7} & 100.0 & 30.6 & 37.2 & 6.9 & 29.6 & 7.0 & \textbf{19.8} & 100.0  \\
\bottomrule[1pt]
\end{tabular}
}
\caption{ Results on the MHQA benchmark by the gpt-3.5-turbo-1106 and Qwen-72B with zero-shot in setting 2. Ans means answer. Sup means supporting paragraph index and tile. Joint means evidence triples including relationship with sub-answers. FSM2 means LLMs summary with results of FSM1 again }
\label{tab:setting2}
\end{table*}

\section{Experiments}
\subsection{Benchmark and Evaluation}
We evaluate our model on three high-quality MHQA datasets: HotpotQA \cite{hotpot}, 2WikiMultiHopQA \cite{2wiki} and Musique \cite{musi}. Learning from the shortcut phenomenon \cite{Min_no_need_mhreason} of single hop questions in HotpotQA, Musique strictly controls the composition of the question, ensuring that it must undergo multiple inferences to find the answer. Both HotpotQA and 2Wiki have ten candidate paragraphs for each question and originally have supporting facts. While Musique has twenty candidate with longer text and no supporting facts. Therefore, Musique is the most standard and difficult MHQA datasets. Following traditions \cite{self-prompted}, We adopt the exact match (EM) and F1 scores as evaluation metrics and conduct experiments on subsets of the datasets by randomly selecting 1000 samples from the test sets. Despite having similar basic instructions and a clearly defined output format for all methods, the model's consistency in following instructions may vary across different methods. This variation can result difficulty for answer extraction during evaluation. To address this issue, we introduce a new metric, format, measuring the accuracy of the output format.
\begin{table}
\centering
\footnotesize
\setlength{\tabcolsep}{4pt}
\begin{tabular}{l l rrrrrrrrr} 
\toprule[1pt]
 & & \multicolumn{2}{c}{Musique} & \multicolumn{2}{c}{HotpotQA} & \multicolumn{2}{c}{2Wiki} \\	\cmidrule(lr){3-4}\cmidrule(lr){5-6}\cmidrule(lr){7-8} 
 
 &  & EM  & F1   & EM   & F1   & EM   & F1  \\ \midrule
 
\multirow{5}{*}{\rotatebox[origin=c]{90}{GPT}} & \textbf{Normal} &  19.2 & 33.3 & 31.9 & 43.7 & 36.0 & 46.6    \\
&\textbf{COT} & 20.6 & 35.6 & 32.1 & 45.5 & 38.1 & \textbf{53.0 }  \\
&\textbf{SP-COT} & 14.4 & 28.4  & 24.8 & 37.4 & 23.2 & 36.0   \\

&\textbf{FSM1} & 23.1 & 40.3 & 24.5 & 39.3 & 27.1 & 40.6 \\
&\textbf{FSM2}  & \textbf{26.7} & \textbf{40.5} & \textbf{33.3} & \textbf{45.7} & \textbf{39.2} & 50.1 \\
 \midrule
\multirow{5}{*}{\rotatebox[origin=c]{90}{Qwen}} & \textbf{Normal} & 12.9 & 19.9  & 31.0 & 41.6 & 31.9 & 39.1   \\
&\textbf{COT} & 14.1 & 24.0  & 30.6 & \textbf{42.7} & 39.9 & 49.8   \\
&\textbf{SP-COT}  & 6.0 & 14.7 & 14.6 & 28.6 & 18.5 & 31.8 \\

&\textbf{FSM1}  & \textbf{33.2}  & \textbf{48.5} & 28.0 &  37.4 & 39.1  & 47.9   \\
&\textbf{FSM2}  & \textbf{33.2}  & \textbf{48.5} &  \textbf{32.2} & 41.3  &  \textbf{40.2} & \textbf{50.3 }  \\

 \bottomrule[1pt]
\end{tabular}

\caption{ Results on the MHQA benchmark by the gpt-3.5-turbo-1106 and Qwen-72B in setting 1. }
\label{tab:setting1}
\end{table}

\subsection{Baselines}
Baseline methods in the experiment:

\textbullet~The \textbf{Normal} is the basic form, involving only task descriptions and output requirements, without explicit instructions for reasoning.

\textbullet~The \textbf{COT} \cite{cot} is widely used in LLMs for inference due to its simplicity and effectiveness. It prompts LLMs to create intermediate step-by-step rationales, aiding in the reasoning process for obtaining answers.

\textbullet~The \textbf{SP-COT} \cite{self-prompted} introduces a pipeline for generating high-quality Open-Domain Multi-step Reasoning (ODMR) datasets. It utilizes an adaptive sampler for case selection and self-prompted inference via ICL. This technique organizes reasoning chains into six categories, inspired by the construction of the Musique \cite{musi} dataset.

\subsection{Setting}
Our study explores two settings: (1) generating answers directly from the context and question, and (2) building a complete reasoning chain that includes the answer, supporting evidence, and facts to assess the coherence of the reasoning process. Due to the lack of gold evidence for Setting 2 in Musique, our evaluation can not evaluate on it.

\subsection{Models}
For MHQA task, we require models with the ability for processing lengthy text. FSM operates in multiple rounds, demanding models capable of handling conversational contexts. We selected GPT-3.5-turbo-32k and Qwen72B-chat \cite{qwen} for our study. Additionally, we employed vllm \cite{vllm} to accelerate the inference process.

\subsection{Results}
The results of setting 1(sole answer) are detailed in  Table \ref{tab:setting1}, while the outcomes for setting 2(answer paired with supporting fact) are displayed in Table \ref{tab:setting2}. 
Our approach demonstrates superior results in setting 2, particularly on the most difficult dataset. This is attributed to the increased complexity of instructions in Setting 2, making it harder for models to follow them accurately. Furthermore, the presence of straightforward single-hop questions in the HotpotQA and 2Wiki datasets \cite{Min_no_need_mhreason} can confuse the LLMs with multi-hop reasoning. While our method's performance in Setting 1 on less complex datasets like HotpotQA and 2Wiki is moderately satisfactory, it excels in precision with fewer instances of hallucination.

The performance of COT is notably inadequate, falling considerably below the standard few-shot settings. This discrepancy is mainly due to its failure to provide answers in the required format, detailed in Figure \ref{format_error}, a flaw we attribute to its bad instruction following ability.
Conversely, the normal method struggles with supporting facts but achieves substantially higher scores on answers. This phenomenon indicates that although LLMs may misinterpret intermediate reasoning steps, they still yield correct answers, hinting at underlying data leakage and speculating. While some errors may stem from misinterpreting instructions, it is evident that there are significant concerns surrounding the authenticity and logical coherence of the models' reasoning chains. Additionally, the prospect of dataset leakage during evaluation cannot be disregarded.
 In conclusion, we posit that our method maintains a competitive edge in this context.

 
\section{Discussion} 
\label{Sec:Disscusion}
Figure.~\ref{fig_error} provides error examples in experiments. We conclude four types of errors. \textbf{a)Reasoning Lost Issue}: providing an answer just with the last sub-question, instead of options for the original sentence. \textbf{b)Formatting Error}: The output can not be parsed to get answer, which added difficulty to the evaluation. Examples are presented in Figure \ref{format_error}.
 \textbf{c)Sub-question Decomposition Error}: Incorrectly decomposed the sub-questions. \textbf{d)Sub-answer Error}: Identified the wrong paragraph but provided a correct answer. \textbf{e)Hallucination Response}: Provided an correct answer without locating the relevant paragraph. 

\section{Conclusion}
We have identified issues in traditional methods where LLMs may produce errors in the intermediate reasoning process but still arrive at the correct answer. Additionally, these methods often require few-shot demonstrations, which may surpass the maximum context length of LLMs. Therefore, we propose an easy zero-shot prompt paradigm called the FSM to address MHQA tasks systematically in an automated format. Our framework approaches problem-solving by focusing on one sub-task at a time iteratively, revising each step to ensure precision. By guiding LLMs through problems incrementally, FSM achieves superior results and aids in enhancing the LLMs' capabilities without resorting to shortcuts.

\section*{Limitations}
This multi-turn dialogue process, inherent to our framework, mandates repeated handling of improperly formatted outputs, due to the output before will be the next input, which can be challenging for models with smaller parameter sizes and weaker follow-instruction capabilities. Therefore, models with limited capacity to follow instructions might not benefit from our method as any error in the intermediate steps could lead to an abrupt termination of the process.

\bibliography{custom}
\bibliographystyle{acl_natbib}

\appendix
\section*{Appendix}
\label{sec:appendix}

\section{Related Work}


\textbf{Multi-hop Question Answering} Existing approaches to solving the multi-hop QA task can be mainly categorized into  question decomposition \cite{decompose_trained,decomposing-complex,decompse_unsupervise}, graph-based method \cite{gnn-hgn,gnnidentifying,gnnmhqa}, iterative method \cite{iterly_q} and LLMs \cite{self-prompted} prompts. These models grapple with computational complexity and extensibility, and they lack an interpretable reasoning chain, which deviates from human cognitive processes.\
\textbf{Language model for reasoning.} CoT\cite{cot} reveals the ability of large language models to formulate their reasoning procedure for problem-solving. Several follow-up works have since been performed, including the least-to-most prompting technique \cite{least_to_most} for solving complicated tasks, zero-shot CoT \cite{zero-cot}, graph-of-thought (GoT) \cite{got}, and reasoning with self-consistency \cite{self-consistency}. ReAct \cite{react} interleaves the generation of reasoning traces with task-specific actions, promoting greater synergy.

\textbf{Task decomposition.} \cite{decompose_trained} decomposes a multi-hop question into a number of independent single-hop sub-questions, which are answered by an off-the-shelf question-answering (QA) model. These answers are then aggregated to form the final answer. Both question decomposition and answer aggregation require training models. After the emergence of Large Language Models (LLMs), traditional training methods \cite{train_kuaishou} are rarely used due to their expensive nature. Most current research focuses on the few-shot approach. \cite{least_to_most} chains the processes of problem decomposition and sub-problem solving. The original problem and its sub-problems are inherently interrelated, and forcibly breaking them down into unrelated problems would unnecessarily increase the difficulty.

\section{Prompt}
\label{prompt}
\subsection{FSM1}
Decomposer = Please determine whether the question is simple sentence or compound sentence. If it is a simple sentence, return {"simple":true,"subquestion":null}.Otherwise, simple: false, decompose the question and generate the first answerable simple sentence. reply in the form of {"simple":false,"subquestion":xxx}. Do not reply any other words and provide answers in JSON format!

Searcher = Given the paragraph below, please find out the paragraph that contains the answer of "{}" Please take a moment to thoroughly understand the content before proceeding to the questions, then carefully read the relevant paragraphs based on the question and provide the most likely answer. Return the title of the paragraph and the answer no more than 5 words in the form of {"question":xxx, "paragraph title":xxx, "answer":xxx}. Do not reply any other words and provide answers in JSON format!

Judge-if-continue= Please compare the complex question and subquestion, answer whether they are semanically identical in the form of {"identical":true or false}. Do not reply any other words and provide answers in JSON format!

\subsection{FSM2}
FSM2-post-summary-again= Documents:
paragraphs:{paragraphs found in FSM1}
subquestion and answers:{subquestion and answers given in FSM1}
Question:{origin question}
Answer the question reasoning step-by-step based on the Doucments. If it is a general question, please respond with 'Yes' or 'No'. Finally, you must return the title of the context, the sentence index (start from 0) of the paragraph and the concise answer no more than 10 words and explaination in the form of {"supporting-facts": [[title, sentence id], ...], "evidences": [[subject entity, relation, object entity],...], "answer":"xxx","explain":"xxxx"}. Do not reply any other words. 

\subsection{Baseline}
SP-COT\cite{self-prompted}= This is a two-hop to four-hop reasoning question-answering task that requires decomposing the questions into simple, answerable single-hop questions. The decomposition process involves four types of questions: comparison, inference, compositional, and bridge-comparison. There are six specific decomposition steps in total, denoted by Q representing the decomposed subproblems. The steps are as follows:
First, Q1 -> Q2
Second, Q1 -> Q2 -> Q3
Third, Q1 -> Q2 -> Q3
Fourth, (Q1\&Q2) -> Q3
Fifth, (Q1\&Q2) -> Q3; Q3 -> Q4
Sixth, Q1 -> Q2; (Q2\&Q3) -> Q4
The process involves first determining the type of question and then identifying the decomposition process type. It's important to note that the decomposition of questions cannot be provided all at once; it must be done step by step. Each subproblem needs to be decomposed and answered before moving on to the next one, as there is interdependence between the subproblems .Finally, you must return the title of the context, the sentence index (start from 0) of the paragraph and the concise answer and explaination in the form of 
{"explain":"xxxx","supporting-facts": [[title, sentence id], ...], "evidences": [[subject entity, relation, object entity],...],"answer":"no sentence and no more than 10 words "}. 
Do not reply any other words.

COT-setting1-w/o-evidence = Answer the question according to the context,Let's think step by step, and explain your reasoning process. You must return in the form of {"explain":"xxxx","answer":answer}. Do not reply any other words.

normal-setting1-w/o-evidence =  Answer the question according to the context. You must return in the form of {"explain":"xxxx","answer":answer}. Do not reply any other words.

normal-setting2-w-evidence =  Answer the question according to the context. Find the paragraph that contains the answer of question, and summarize a triple that contains [subject entity, relation, object entity]. Finally, you must return the title of the context, the sentence index (start from 0) of the paragraph and the concise answer no more than 10 words in the form of {"supporting-facts": [[title, sentence id], ...], "evidences": [[subject entity, relation, object entity],...], "answer":answer}. Do not reply any other words.

prompt-step =  Answer the question according to the context,Let's think step by step, and explain your reasoning process. Find the paragraph that contains the answer of question, and summarize a triple that contains [subject entity, relation, object entity]. Finally, you must return the title of the context, the sentence index (start from 0) of the paragraph and the concise answer no more than 10 words in the form of {"supporting-facts": [[title, sentence id], ...], "evidences": [[subject entity, relation, object entity],...], "answer":answer}. Do not reply any other words.

React-setting2-w-evidence =  Solve a question answering task with interleaving Thought, Action, Observation steps. Thought can reason about the current situation, and Action can be three types: 
(1) Search[entity], which searches the exact entity on given context and returns the first paragraph if it exists. If not, it will return some similar entities to search.
(2) Lookup[keyword], which returns the next sentence containing keyword in the current passage.
(3) Finish[results], which returns the answer and finishes the task.
You should plan and reason in the Thought, then perform your Action, lastly, observe the result of action. Loop this process until the problem was finished. 
At last, you must additional output the title of the paragraphs, the sentence index (start from 0) of the paragraph and the concise answer no more than 10 words and explaination in the form of 
Thought: reasoning
Action: Search[entity] or Lookup[keyword] or Finish[results]
Observation: observe the results of action
end with Finish[{"supporting-facts": [[title, sentence id], ...], "evidences": [[subject entity, relation, object entity],...], "answer":answer}]

\subsection{Format Error}
\noindent\begin{minipage}{\textwidth}
  \begin{figure}[H]
    \includegraphics[width=\textwidth]{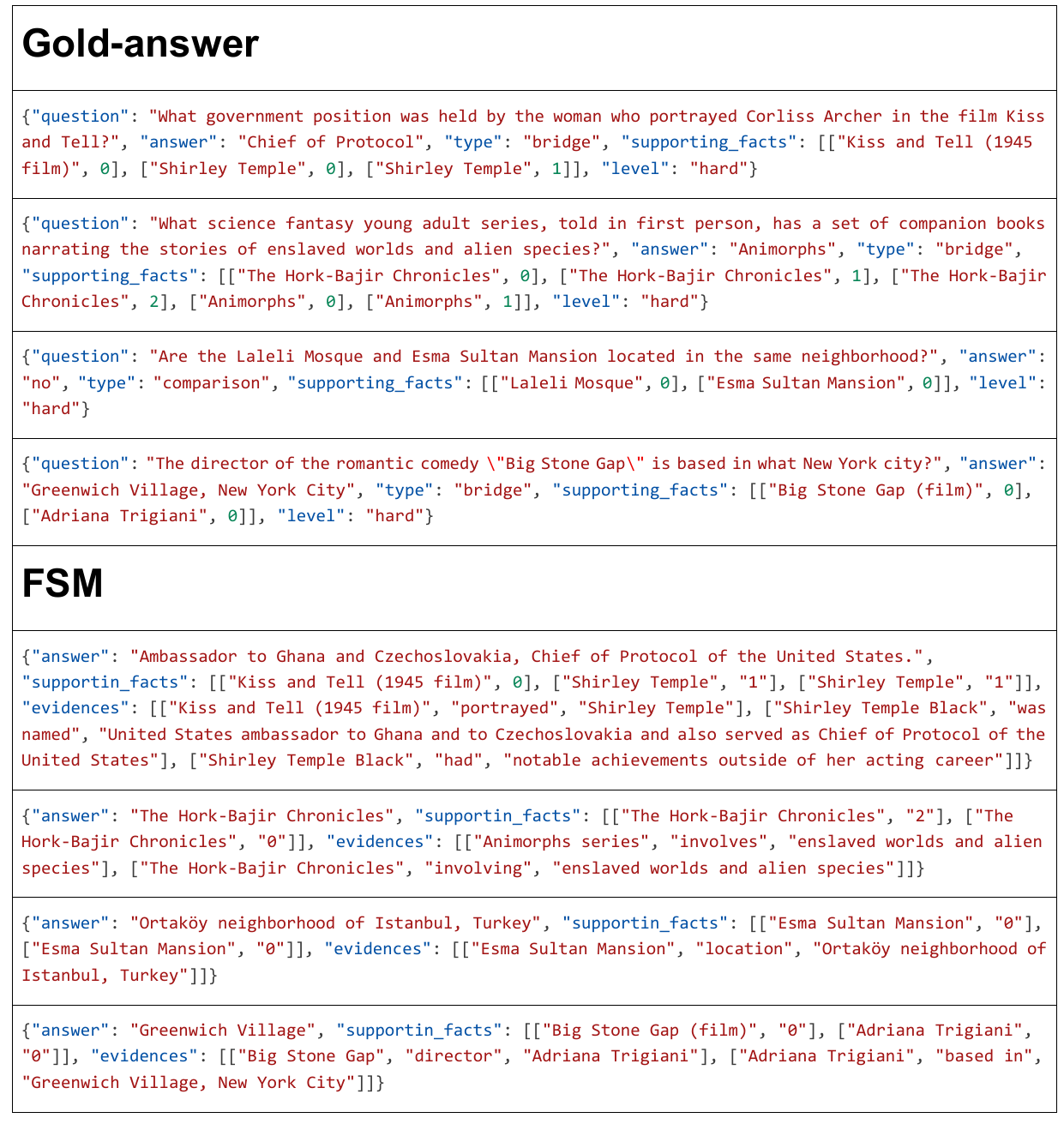}
    \caption{The outputs of FSM are standard json format.}
    \label{format_error1}
  \end{figure}
\end{minipage}

\clearpage

\noindent\begin{minipage}{\textwidth}
  \begin{figure}[H]
    \includegraphics[width=\textwidth]{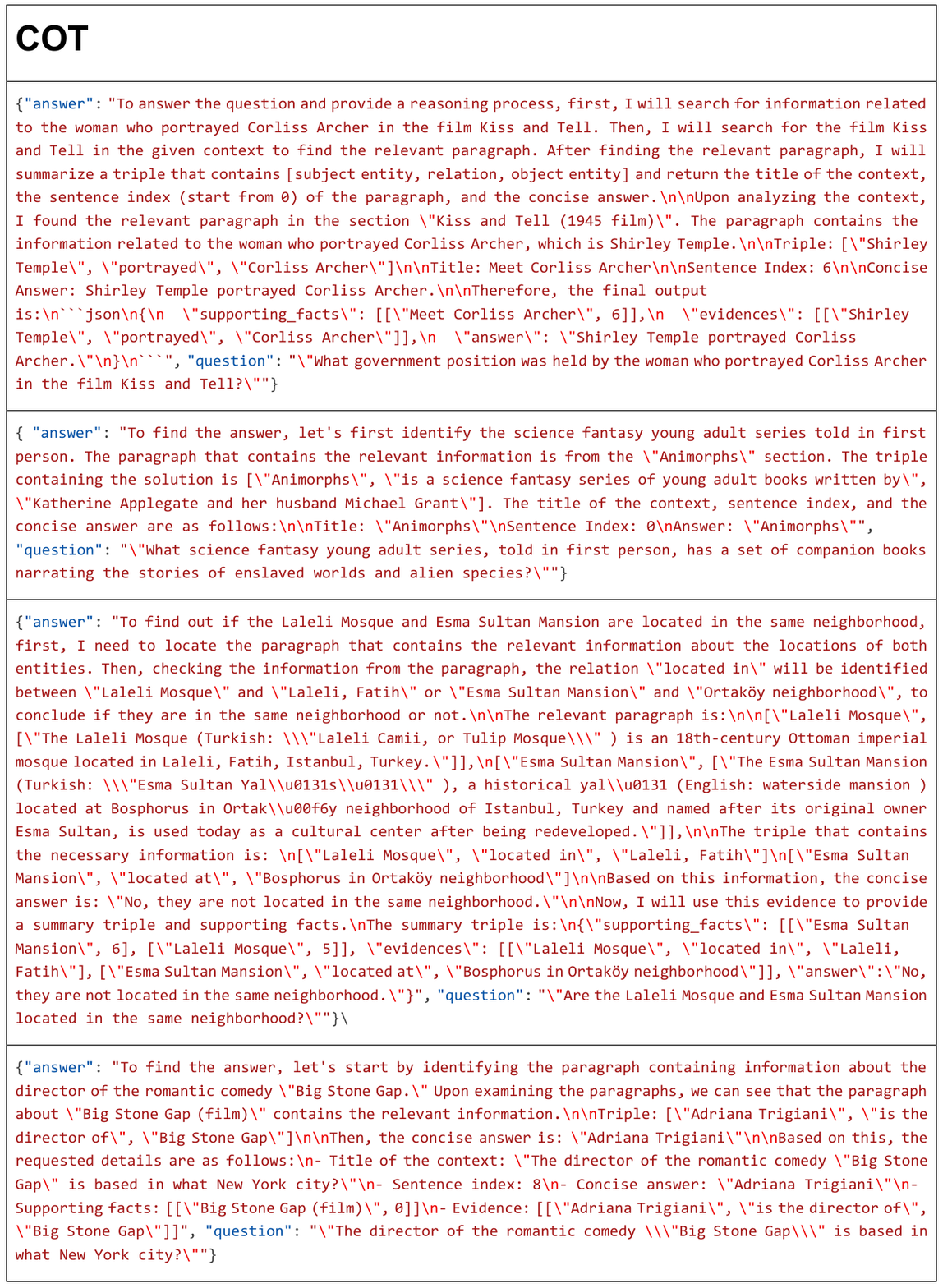}
    \caption{There are some error format examples for COT.}
    \label{format_error}
  \end{figure}
\end{minipage}

\begin{figure*}[ht!]
  \centering
  \includegraphics[width=\textwidth]{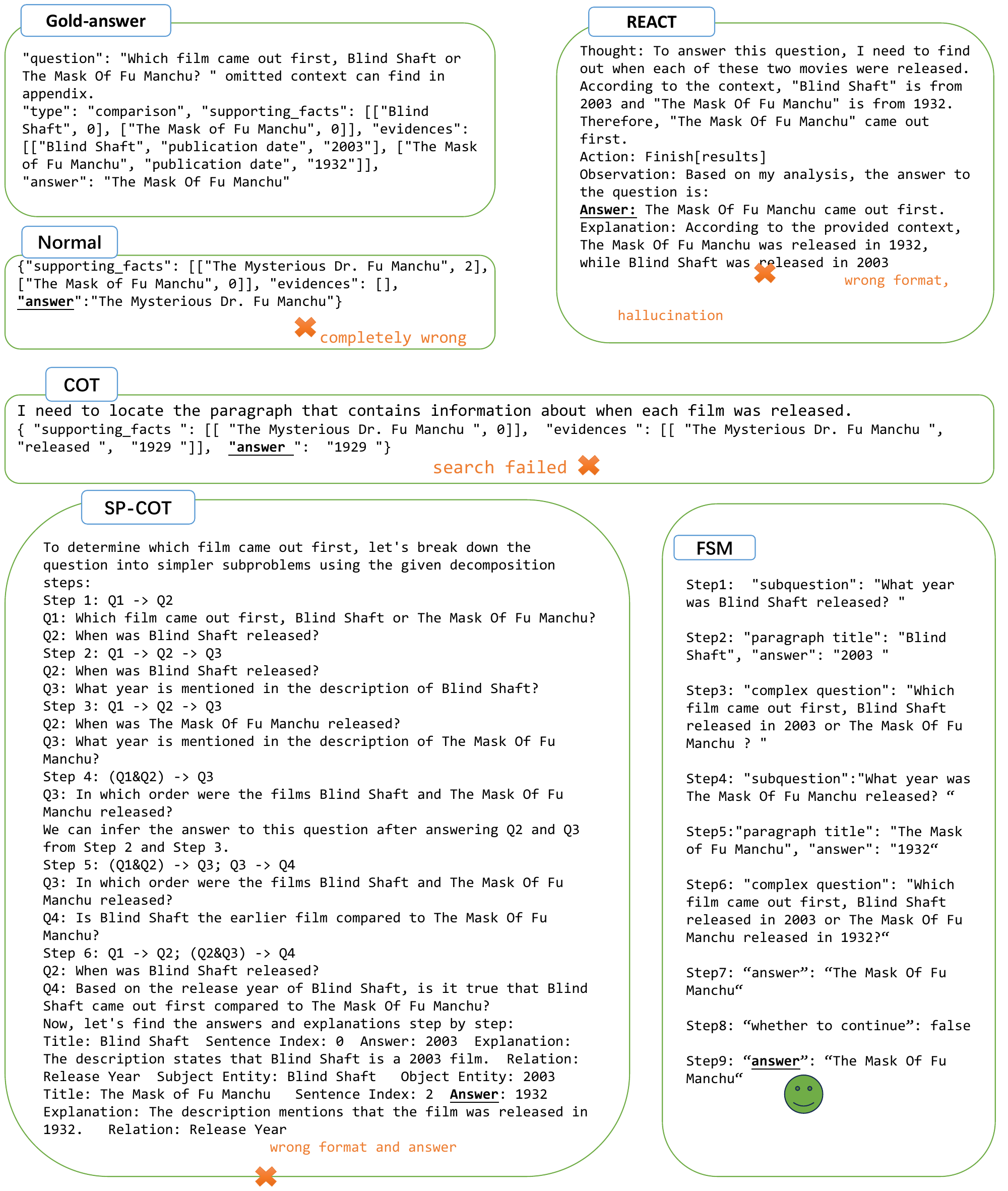}
  \caption{Contrast between baseline and FSM. There are some error examples for baseline.}
  \label{fig_error}
\end{figure*}
\end{document}